\documentclass[10pt,twocolumn,letterpaper]{article}

\usepackage{cvpr}              

\usepackage[accsupp]{axessibility}  

\usepackage{graphicx}
\usepackage{amsmath}
\usepackage{amssymb}
\usepackage{booktabs}

\usepackage{multirow}
\usepackage{booktabs}
\usepackage{boldline,makecell}
\usepackage{subcaption}
\usepackage{threeparttable}

\usepackage{scalerel}
\usepackage[dvipsnames]{xcolor}
\definecolor{lightblue}{RGB}{153, 207, 224}
\definecolor{seagreen}{RGB}{0, 210, 210}
\definecolor{coral}{RGB}{247,100,100}
\definecolor{grey}{RGB}{153,141,141}
\newcommand{\UI}{\scaleto{UI}{3.2pt}}

\usepackage[pagebackref,breaklinks,colorlinks]{hyperref}

\usepackage[capitalize]{cleveref}
\crefname{section}{Sec.}{Secs.}
\Crefname{section}{Section}{Sections}
\Crefname{table}{Table}{Tables}

\def\cvprPaperID{2275} 
\def\confName{CVPR}
\def\confYear{2023}

\begin{document}

\title{Recognizability Embedding Enhancement for Very Low-Resolution Face Recognition and Quality Estimation}

\author{Jacky Chen Long Chai\textsuperscript{1} \ Tiong-Sik Ng\textsuperscript{1} \ Cheng-Yaw Low\textsuperscript{2} \ Jaewoo Park\textsuperscript{1} \ Andrew Beng Jin Teoh\textsuperscript{1}\\ 
\textsuperscript{1}Yonsei University \ \textsuperscript{2}Institute for Basic Science\\
{\tt\small \textsuperscript{1}\{jackyccl,ngtiongsik,julypraise,bjteoh\}@yonsei.ac.kr \ \textsuperscript{2}\{chengyawlow\}@ibs.re.kr}
}

\twocolumn[{
\maketitle
\begin{center}
    \captionsetup{type=figure}
    \includegraphics[width=.99\linewidth]{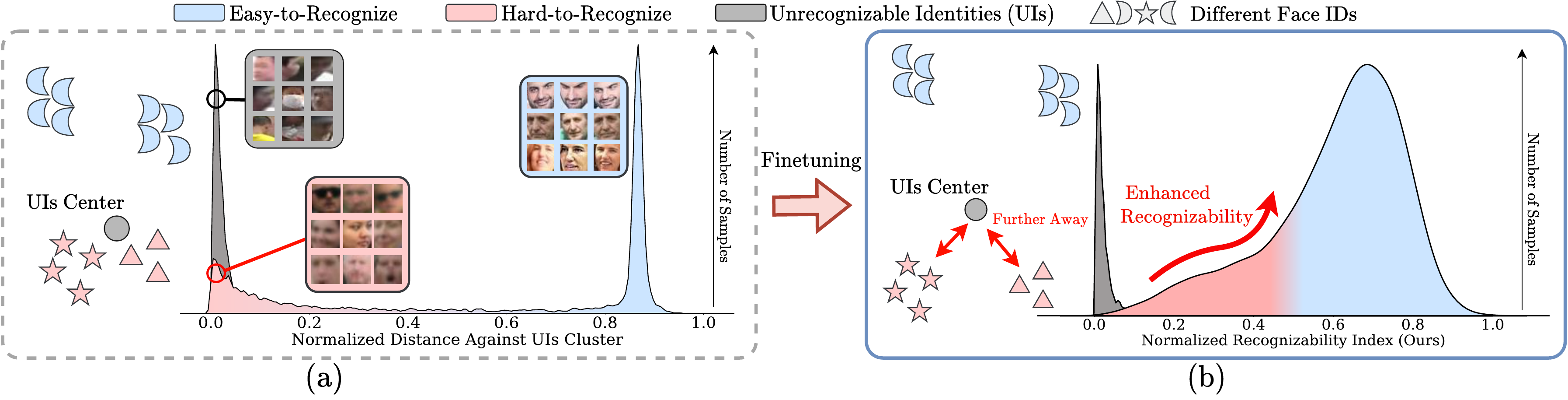}
    \vspace{-3mm}
    \caption{\textcolor{black}{A deep face model pretrained on high-resolution face images introduces a cluster of unrecognizable instances (\textcolor{grey}{\textbf{grey spike}}), dubbed \textit{unrecognizable identities} (UIs) in \cite{deng2021harnessing}. (\textbf{a}) shows the bimodal distribution for a very low-resolution face dataset \cite{cheng2018low_tinyface} based on distance against the UIs. Interestingly, a portion of hard-to-recognize faces (\textcolor{coral}{\textbf{red peak}}) lie close to the UIs, indicating their low recognizability. (\textbf{b}) We propose to improve the recognizability of hard-to-recognize instances by pushing them away from the UIs center. Consequently, faces with higher recognizability indexes are further apart from UIs center in the embedding space. Our method not only induces more discriminative representations but also translates face quality into a measurable indicator that closely matches human cognition.}}

    \label{fig::away_from_unrecog_faces_overall_ideas}
\end{center}
}]

\begin{abstract}
\vspace{-1mm}
Very low-resolution face recognition (VLRFR) poses unique challenges, such as tiny regions of interest and poor resolution due to extreme standoff distance or wide viewing angle of the acquisition devices. In this paper, we study principled approaches to elevate the \textit{recognizability of a face in the embedding space} instead of the visual quality. We first formulate a robust learning-based face recognizability measure, namely recognizability index (RI), based on two criteria: (i) proximity of each face embedding against the unrecognizable faces cluster center and (ii) closeness of each face embedding against its positive and negative class prototypes. We then devise an index diversion loss to push the hard-to-recognize face embedding with low RI away from unrecognizable faces cluster to boost the RI, which reflects better recognizability. Additionally, a perceptibility attention mechanism is introduced to attend to the most recognizable face regions, which offers better explanatory and discriminative traits for embedding learning. Our proposed model is trained end-to-end and simultaneously serves recognizability-aware embedding learning and face quality estimation. To address VLRFR, our extensive evaluations on three challenging low-resolution datasets and face quality assessment demonstrate the superiority of the proposed model over the state-of-the-art methods.
\end{abstract}
\vspace{-2mm}

\section{Introduction}
\label{sec:intro}

In real-world face recognition deployment scenarios, the pixel resolution of the detected face images is significantly deflated, due to extreme long-range distance and broad viewing angle of the acquisition devices, especially in surveillance applications. These tiny regions of interest are, in general, ranging from 16$\times$16 to 32$\times$32 pixels \cite{zou2011very}, thereby suffering from poor pixel resolution, in addition to unrestricted  noises such as poor illumination conditions, non-frontal poses with awful angles, unconstrained facial expressions, blurriness, and occlusions \cite{terhorst2020ser_fiq}. It is noteworthy that these contaminated very low-resolution (VLR) face images undermine the overall performance of \textcolor{black}{a face model trained with its high-resolution (HR) counterparts; therefore, there is a lack of generalizability to resolve the VLR face recognition (VLRFR) problem \cite{cheng2018low_tinyface}. Apart from that, training of a VLRFR model often suffers} from \textcolor{black}{very limited representative face examples to extract meaningful identity-specific patterns}. These issues are further escalated due to ambiguous inter-class variations for the heavily distorted face instances with perceptually similar identities in particular $ $\cite{singh2021derivenet}. \textcolor{black}{Whilst matching a probe to a gallery set of the same resolution (i.e. VLR to VLR) is still an open challenge, the resolution gap between galleries and probes triggers another problem in cross-resolution matching (typically HR galleries to VLR probes). Hence, the generalization performance of the prevalent deep learning models for VLRFR is still far from satisfactory.} 

\textcolor{black}{As a whole, most existing works designated for VLRFR improve the face quality of the VLR instances based on an auxiliary set of HR face images \cite{li2018face}}. The generic operation modes are either in image domain (\textit{super-resolution, image synthesis}) \cite{yue2016image_super_resolution, zhang2021faceHalluwiffinishingtouches, yin2020FAN}, embedding domain (\textit{resolution-invariant features, coupled mappings}) \cite{lu2018deepcoupled,talreja2019coupledgan}, or at classifier level (\textit{transfer learning, knowledge distillation}) \cite{ge2018lowwildknowledgedistillation, ge2020lookoneandmore,huang2020DDL, shi2020URL}. However, most of these state-of-the-art models require mated HR-VLR pairs of the same subject. This is unrealistic in practice as the HR-VLR pairs are often unavailable. 

As far as face recognition is concerned, face recognizability (\textit{also known as face quality} \cite{hernandez2020biometric_faceqnet, hernandez2019faceqnet}) can be deemed as a \textit{utility} of how well a face image is for discrimination purposes. In other words, face quality is closely related to face recognition performance. Some works thus focus on predicting a face image's suitability for face recognition, commonly known as Face Image Quality Assessment (FIQA) \cite{best2018facequalityassessanilk, hernandez2019faceqnet}. FIQA focuses either on (\textbf{i}) creating propositions to label the training data with face image quality scores and solve a regression problem \cite{ou2021sdd_fiqa, hernandez2019faceqnet, hernandez2020biometric_faceqnet}, or (\textbf{ii}) linking the face embedding properties to FIQ scores \cite{chang2020DUL, terhorst2020ser_fiq, shi2019pfe, meng2021magface, kim2022adaface}. The second approach shows better quality estimation, with the possible reason that the first approach is prone to mislabeling of ground truth quality \cite{meng2021magface, terhorst2020ser_fiq}. However, the second approach may not be optimal since the FIQ scores are estimated based on the embedding properties rather than through a learning process \cite{boutros2021crfiqa}. 

\textcolor{black}{Recently, \cite{deng2021harnessing} reported an intriguing observation that a deep learning-based face model induces an unrecognizable cluster in the embedding space. The cluster, known as \textit{unrecognizable identities} (UIs), is formed by unrecognizable face examples, owing to diverse inferior quality factors, including VLR, motion blurred, poor illumination, occlusion, etc. Hence, these face examples with varying ground truths incline to lie close to the UIs, rather than their respective identity clusters. This observation inspires us to analyze the embedding distribution of the VLR face images against the UIs center. Interestingly, the extreme bimodal distribution in \cref{fig::away_from_unrecog_faces_overall_ideas} discloses that a significant number of the VLR faces in TinyFace \cite{cheng2018low_tinyface}, i.e., a realistic VLR face dataset, are \textit{hard-to-recognize} from the human perspective and therefore rendered next to the UIs cluster. We reckon that mining representative patterns from these hard-to-recognize faces is more meaningful for face recognition, in place of defining them as the elements of UIs. Apart from that, a more reliable objective quality metric is needed to better interpret each VLR face example in terms of its embedding recognizability for recognizability-aware embedding learning.}

Instead of perceptual quality, this work aims to elevate the recognizability of every VLR face embedding. In a nutshell, we formulate a learning-based recognizability index (RI) with respect to the \textit{Cosine proximity} of each embedding instance with (i) the UIs cluster, and (ii) the associated positive and negative prototypes. In the meantime, the index diversion (ID) loss is presented to detach the hard-to-recognize embeddings from the UIs cluster, alongside a perceptibility attention mechanism. We underline that embedding learning in the direction opposing the UIs contributes to a higher explanatory power whilst promoting inter-class separation, particularly for hard-to-recognize instances. 

For clarity, we summarize our contributions as follows:
\begin{itemize}
\item A novel approach is proposed to address the VLRFR, including VLR-VLR and HR-VLR matching, by leveraging the \textit{face recognizability} notion in the embedding space to improve the hard-to-recognize instances.
\item A robust learning-based face recognizability, dubbed RI, is put forward. RI relies on the face embeddings' intrinsic proximity relationship against the UIs cluster, positive, and negative class prototypes.
\item An index diversion (ID) loss is devised to enhance the RI for face embeddings. Further, we put forward a perceptibility attention mechanism to guide embedding learning from the most salient face regions.
\item Our proposed model trained in an end-to-end manner not only renders a more discriminative embedding space for VLRFR but simultaneously serves recognizability-aware embedding learning and face recognizability estimation.
\end{itemize}

\section{Related Work}
\label{sec:related_work}
\textbf{Very Low-Resolution Face Recognition.} Existing HR image dependence approaches for VLRFR can be categorized into image domain, embedding domain, and classifier. Under the image domain, the super-resolution (SR) model learns a mapping function to upscale VLR into HR images to improve faces' visual quality \cite{yue2016image_super_resolution, singh2018identityawaresynthesis}. Several successors' works \cite{hsu2019sigan, yu2019semanticfacehallucination, zhang2021faceHalluwiffinishingtouches, yin2020FAN} relate recognition to visual quality. However, these works require mated HR-VLR pairs of the same subject to be available for embedding learning, which is unrealistic.

At both embedding domain and classifier levels, most knowledge distillation (KD) aimed to transfer the knowledge of the HR domain to the VLR face model via a teacher-student network configuration \cite{ge2018lowwildknowledgedistillation}. To achieve cross-resolution distillation, \cite{ge2020lookoneandmore} employed an additional network to bridge the teacher and student network, while  \cite{massoli2020crossresolutionKD} compared the face embeddings of teacher and student networks with a regression loss. On the other hand, distribution distillation loss \cite{huang2020DDL}, non-identity-specific mutual information \cite{low2021mindnet} and implicit identity-extended augmentation \cite{low2022implicit} have also been explored to mitigate the performance gap between HR and VLR instances. \cite{singh2021derivenet} reconstructed an HR embedding to approximate the class variations of VLR instances while learning similar features for HR and VLR images. However, despite their high visual quality, some HR faces may not be recognized. In fact, \cite{zhang2018super_identity, li2019low_resolution_face_recog_in_wild, cheng2018low_tinyface} suggest that improving visual quality can undermine identity-specific traits important to the downstream recognition task. Our proposed method involves no auxiliary HR images. We aim to improve the hard-to-recognize instances based on the recognizability notion rather than visual quality.

Recently, Li \etal \cite{li2022deeprival} proposed a rival margin on the hardest non-target logits to maximize the separation against the nearest negative classes. However, low-quality face images with perceptually similar identities are close to each other. Therefore, enforcing separation for a low-quality face image only from the nearest non-target class hardly learns meaningful characteristics to differentiate the two identities. On the contrary, we strive to enlarge the overall inter-class dissimilarity by enhancing image recognizability.

\textbf{Face Image Quality Assessment (FIQA)} can be mainly divided into two categories. The first category is to solve a regression problem to assess the training images with FIQ scores \cite{best2018facequalityassessanilk, hernandez2019faceqnet, hernandez2020biometric_faceqnet, ou2021sdd_fiqa, xie2020inducing, boutros2021crfiqa}. The FIQ scores include human quality annotation \cite{best2018facequalityassessanilk}, the intra-class Euclidean distance between an instance and an ICAO \cite{wolf2016portrait} compliance instance \cite{hernandez2019faceqnet, hernandez2020biometric_faceqnet}, the similarity between random positive mated pairs \cite{xie2020inducing}, the discriminability on each instance \cite{boutros2021crfiqa} and the Wasserstein distance between intra and inter-class distributions \cite{ou2021sdd_fiqa}. 
The second category directly utilizes the intrinsic properties of face embeddings to estimate face quality without explicit regression learning. For instance, \cite{terhorst2020ser_fiq} defined the robustness of stochastic embeddings as FIQ score. However, the computational cost is significant as the assessed instance is required to pass through the network multiple times at different dropout rates. \cite{shi2019pfe} and \cite{meng2021magface} relied on uncertainty variation and embedding norm of the face embeddings as the FIQ scores, respectively. 

Aside from face quality, several works also explore classifiability according to face quality. Instead of using the fixed Gaussian mean as the face embedding \cite{shi2019pfe}, \cite{chang2020DUL} proposed to learn a Gaussian mean alongside an uncertainty to reduce the adverse effects of noisy samples. In \cite{huang2020curricularface}, the easy instances are first learned before the hard ones based on the adaptive margin angular loss. \cite{wang2020mis} defined hard samples and increased the weight of their negative cosine similarities with a preset constant. \cite{meng2021magface} assigned margins based on the embeddings' norm. As a successor, Kim \etal \cite{kim2022adaface} refined the decision on adaptive margin, which only emphasizes hard instances when necessary.

While most FIQA methods struggle to learn adaptive margins for competent quality-aware embedding \cite{boutros2021crfiqa}, we strive to improve embedding learning based on the proposed RI. We demonstrate that RI characterizes the image quality better, and our model can extract meaningful semantics important to face recognition in \cref{sec:experiments_and_results}.

\begin{figure}[t]
  \centering
   \includegraphics[width=0.99\linewidth]{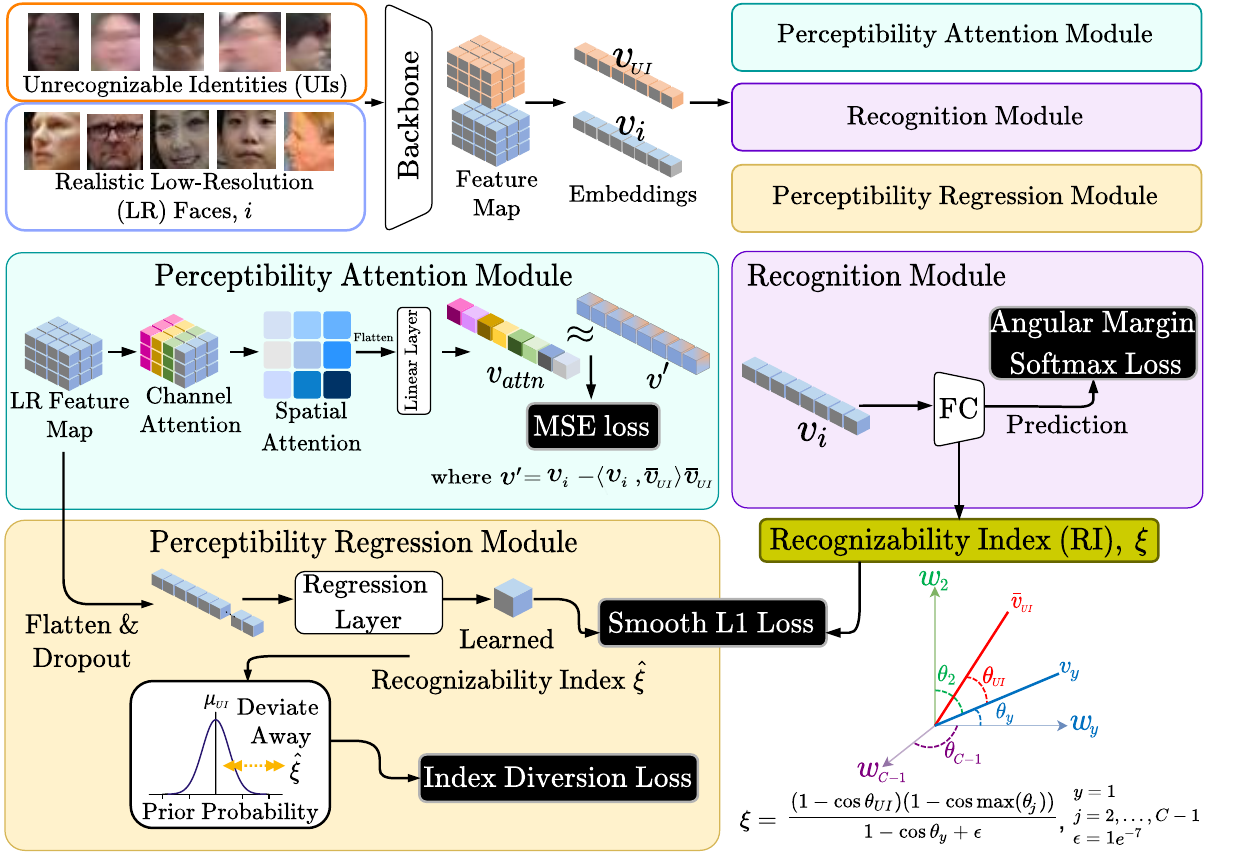}
   \vspace{-3mm}
    \caption{Our proposed model comprises three main modules: (i) \textit{recognition module} learns the face embeddings ${v_i}$ by optimizing the class prototypes for the recognition task; (ii) \textit{perceptibility regression module} is designed to learn the recognizability index, $\hat{\xi}$ thereby enabling the recognizability prediction for any samples including the unseen ones; and (iii) \textit{perceptibility attention module} performs channel and spatial-wise attention on the embeddings that approximate the projection away from the UIs cluster, ${v'}$. The RI is learned based on two criteria, as shown on the bottom right, where $y$ is the target class, $j$ is the non-target class across all $C$ identities, and ${\bar{v}}_{\UI}$ is the mean of UI cluster embeddings.}
    \label{fig::arch}
    \vspace{-2mm}
\end{figure}

\section{Methodology}
\label{sec:methodology}

\subsection{Recognizability Index (RI) Formulation}
\label{sec:RI}
A face model trains a discriminative embedding space by enforcing intra-class compactness and inter-class discrepancy. 

Interestingly, \cite{ou2021sdd_fiqa} discloses the relationship between face quality and recognition performance by computing the Wasserstein distances between intra-class and inter-class similarity distributions. Hence, it is suggested that face recognizability can be quantified by intra-class and inter-class similarity measures.

In our disposition, the recognizable instances are pushed closer to their positive prototype and further apart from negative prototypes upon convergence. Meanwhile, the hard-to-recognize instances can hardly be pulled toward their positive prototypes, and are usually surrounded by negative ones. Among all negative prototypes, the nearest negative prototype is chosen for inter-class proximity estimation. 

For each instance $i$, the intra-class and the inter-class proximity of its L2-normalized embedding $\widehat{\boldsymbol{v}_{i}}$ with respect to its positive prototype $\boldsymbol{w}_{y_{i}}$ and the nearest negative prototype $\boldsymbol{w}_{j}, j \neq y_{i}$ across a set of $C$ identities are as follows:
\begin{equation}\label{eq::intra_inter_cosine_dist}
    d_{i}^{P} = 1 - \cos(\theta_{y_{i}}) \;,\;
    d_{i}^{N} = 1 - \underset{j \in \{1, \dots, C \} \setminus \{ y_i \}}{\max} \cos (\theta_{j})
\end{equation}
where $\theta_{y_i}$ is the positive angle between $\widehat{\boldsymbol{v}}_{i}$ and $\boldsymbol{w}_{y_i}$ and 
$\theta_{j}$ is the negative angle between $\widehat{\boldsymbol{v}}_{i}$ and $\boldsymbol{w}_{j}$. On the other hand, feeding the unrecognizable faces to the model induces a UIs cluster in the embedding space \cite{deng2021harnessing}. Given the UIs cluster center, i.e.,  the average across all normalized UIs embeddings, $\overline{\boldsymbol{v}}_{\UI}$, the proximity between $\widehat{\boldsymbol{v}}_{i}$ and $\overline{\boldsymbol{v}}_{\UI}$ is defined as:
\vskip -0.5em
\begin{equation}\label{eq::UI_cosine_dist}
    d_{i}^{\UI} = 1 - \cos (\theta_{\UI_{i}})
\end{equation}
where $\theta_{\UI_{i}}$ is the angle between $\widehat{\boldsymbol{v}}_i$ and $\overline{\boldsymbol{v}}_{\UI}$. Since the proximity is in terms of Cosine distance, the instances closest to the UIs cluster center (computed with the smallest $d_{i}^{\UI}$) are referred to as hard-to-recognize instances, and vice versa. Since the proximity of each instance differs with respect to its positive and negative prototypes alongside the UIs cluster center, the association of \cref{eq::intra_inter_cosine_dist} and \cref{eq::UI_cosine_dist} can serve to estimate the face quality. Given an imposed $\epsilon=1e^{-7}$ to avoid division by zero, we define a recognizability measure, dubbed \textit{Recognizability Index} (RI), $\xi_i$ as follows:
\begin{equation}\label{eq::RI}
    \xi_{i} = 
    d_{i}^{\UI} \frac{d_{i}^{N}}{d_{i}^{P} + \epsilon}
\end{equation}

\subsection{Perceptibility Regression Module}
\label{sec:PRM}
As our model serves recognizability-aware embedding learning and quality estimation simultaneously, we introduce a perceptibility regression module (yellow) as in \cref{fig::arch}. The input to the regression module is a flattened feature map learned from the backbone network, navigating through a dropout layer and a fully-connected regression layer to yield a learnable RI, denoted by $\widehat{\xi}_i$. To match $\widehat{\xi}_i$ to $\xi_i$ in \cref{eq::RI}, we apply the smooth L1 loss \cite{girshick2015fastrcnn_smoothL1_loss} as follows:
\begin{equation}\label{eq::smoothL1}
L_{L1}
=
\begin{cases}
0.5(\xi_i - \widehat{\xi}_i)^2 / \beta & \text{if } \lvert \xi_i - \widehat{\xi}_i \rvert < \beta \\
| \xi_i - \widehat{\xi}_i | - 0.5\times\beta & \text{otherwise}
\end{cases}
\end{equation}
where $\beta$ is a threshold that switches between L1 and L2-losses. In the early training stage, the L1 loss computes consistent gradients to approximate $\widehat\xi$ towards $\xi$. When the regression error falls below the confidence interval, the L2 loss exhibits a smoother transition in the loss surface to facilitate convergence. In our experiments, we fix $\beta=0.75$. It is noteworthy that the RI from the regression module differs from ERS \cite{deng2021harnessing} in the following perspectives: (i) ERS does not consider the intra-class and inter-class proximities, and (ii) ERS requires the UIs cluster center to be known for recognizability score estimation, but our model simply withdraws the UIs cluster during the deployment. In other words, upon training completion, the regression module incorporates the proximity relation in \cref{eq::RI} and allows the face quality prediction on any unknown samples without involving the UIs cluster.

\subsection{Index Diversion Loss}
\label{sec:ID_loss}
Having RI obtained from the perceptibility regression module, the next following goal is to enhance the hard-to-recognize instances' recognizability. We first model the RI distribution of the UIs cluster. According to the Central Limit Theorem \cite{fischer2011historyofCLT}, the Gaussian distribution is the most general distribution for fitting values derived from Gaussian or non-Gaussian random variables. Motivated by this, we can either estimate the mean and the variance of the Gaussian by forwarding the UIs through the perceptibility regression module or simply assume RI follows the standard Gaussian distribution with $\mathcal{N}(0, 1)$. We opt for the latter as it has better interpretability and a more stable RI to resolve the hard-to-recognize instances.

Thus, we define the diversion of the estimated $\widehat{\xi}_i$ as:
\vskip -1.0em
\begin{equation}\label{eq::deviation}
    div = \frac{\widehat{\xi}_i - \mu_{\UI}}{\sigma_{\UI}}
\end{equation}
Since the range of $\widehat{\xi}_i$ is arbitrary, it is rescaled with respect to $\mu_{\UI}$ and $\sigma_{\UI}$ of the UI distribution as:
\vskip -1.5em
\begin{equation}\label{eq::mean_std_ref}
\mu_{\UI} = \frac{1}{K} \underset{k=1}{\overset{K}{\sum}} s_{k} \;, \; \sigma_{\UI} = \sqrt{\frac{\sum_{k=1}^{K}(s_k - \mu_{\UI})^2}{K-1}}
\end{equation}
\vskip -0.5em
\noindent
where $s_{k}$ denotes the RI of a random UI instance that is assumed to be i.i.d and follows $\mathcal{N}(\mu, \sigma^2)$. Here, we set $K=5,000$.
In accordance with \cref{eq::deviation}, the index diversion (ID) loss is formulated as:
\vskip -1em
\begin{equation}\label{eq::id_loss}
    L_{ID} = \:\text{max}(0, \tau - div)
\end{equation}
\noindent
where $\tau$ is the confidence interval hyperparameter. The ID loss enforces a deviation of at least $\tau$ between $\widehat{\xi}_i$ and $\mu_{\UI}$. As a hard-to-recognize instance is associated with a small $\widehat{\xi}_i$, it induces a relatively large ID loss. We attempt to push the hard-to-recognize instances outside the designated $\tau$ by minimizing the ID loss. In other words, the ID loss is equivalent to enforcing a statistically significant deviation of $\mu_{\UI}$ from the estimated $\widehat{\xi}_i$ of VLR instances in the upper tail. Note that although the ID loss enforces a confidence interval, it does not divert $\widehat{\xi}_i$ infinitely as the $\widehat{\xi}_i$ is still bounded to RI under the constraint of \cref{eq::smoothL1}. 

Since the ID loss is dedicated to differentiable learning of the recognizability measure, we argue that the statistical diversion of $\widehat{\xi}_i$ contributes to a more meaningful enhancement of recognizability in the embedding space, especially for the hard-to-recognize face images.

\subsection{Perceptibility Attention Module}
\label{sec:PAM}
In \cite{deng2021harnessing}, Deng \etal showed that diverting the embedding's direction from the UIs cluster center results in face recognition improvement in the inference stage. Given the embedding vector of a VLR face instance $\boldsymbol{v}_i$ and the L2-normalized UIs cluster center $\overline{\boldsymbol{v}}_{\UI}$, we define the embedding projected in the direction away from the UIs cluster $\boldsymbol{v}'_{i}$ based on \cite{deng2021harnessing} as:
\begin{equation}\label{eq::embedding_away_from_UI}
\boldsymbol{v}'_{i} = 
\boldsymbol{v}_i - \langle \boldsymbol{v}_i, \overline{\boldsymbol{v}}_{\UI} \rangle \overline{\boldsymbol{v}}_{\UI}
\end{equation}

We conjecture that the embedding projection away from the UIs cluster center is beneficial to alleviate the model's inadequacy to highlight the meaningful features when the face is obscure. Different from the conventional attention models for the classification tasks, we seek to explore the most salient feature representation with the greatest significance for recognizability by approximating $\boldsymbol{v}_i'$ through an attention module. Inspired by \cite{woo2018cbam}, we design a perceptibility attention module that attends to the spatial and channel dimension of the instances' feature map sequentially. The attended feature maps are fed into a linear fully connected layer to learn an attended embedding $\boldsymbol{v}^{attn}_{i}$ at the same dimension as $\boldsymbol{v}'_{i}$. Lastly, we utilize the mean squared error loss and formalize the regression problem as:
\vskip -1em
\begin{equation}\label{eq::mse}
L_{MSE} = \frac{1}{B}\sum_{i=1}^{B}(\boldsymbol{v}'_i-\boldsymbol{v}^{attn}_{i})^2 
\end{equation}
where $B$ is the batch size. We argue that the projection of embeddings away from the UIs cluster can be deemed as RI-enhanced embeddings. This guides our model to attend to the parts of embedding that contain the richer interpretive contents important to recognition purposes.

Therefore, introducing an attention module permits our model to attend to the most salient face regions, i.e., eyes, nose, etc., of a VLR face instance. In compliance with \cref{eq::smoothL1}, (\ref{eq::id_loss}) and (\ref{eq::mse}), the overall loss function is expressed as follows:
\begin{equation}\label{eq::overall_euqation}
L_{total} = L_{cls} + \alpha L_{L1} + \beta L_{ID} + \gamma L_{MSE}
\end{equation}
where we opt for ArcFace \cite{deng2019arcface} as the classification loss, $L_{cls}$. $\alpha, \beta, \text{and } \gamma$ are the weighting factors for each loss term.

\section{Experiments and Results}
\label{sec:experiments_and_results}
\textbf{Datasets}. To confront the real-world VLRFR problem, our experiments are conducted on three realistic LR face datasets, namely TinyFace \cite{cheng2018low_tinyface}, SurvFace \cite{cheng2018survface}, and SCFace \cite{grgic2011scface} under an open-set evaluation protocol, given the identity labels are disjointed across training and testing sets. We assemble an unlabeled UI face dataset from two public person re-identification repositories by the MTCNN face detector \cite{zhang2016mtcnn}, including LPW \cite{song2018lpwdataset} and MARS \cite{zheng2016mars}. Notably, most of the detected faces are unrecognizable, thereby facilitating the generation of an UIs cluster. The details of each dataset are provided in \cref{sec::dataset_description} of in supplemental material.

\textbf{Experiment Settings}. Our experiments utilize MobileFaceNet \cite{chen2018mobilefacenets} and ResNet-50 \cite{deng2019arcface} pretrained on the VGGFace2 \cite{cao2018vggface2} dataset as the representation encoder. For each dataset, we fine-tune these models using the respective training examples for performance evaluation. Our baseline model is the counterpart trained only with the ArcFace loss. We provide our experimental setup and other relevant settings in \cref{section::experiment_settings} of the supplemental material.

\textbf{Evaluation Metrics}. We summarize the overall performance in rank-1 identification rate (IR) ($\%$) for TinyFace and SCFace. On the contrary, we report the positive identification rate (TPIR) ($\%$) @ false positive identification rate (FPIR), and true positive rate (TPR) ($\%$) @ false acceptance rate (FAR) ($\%$) for SurvFace, due to the inclusion of non-mated face images in its testing set \cite{cheng2018survface}.

To evaluate the characterization of face image quality, we provide the Error versus Reject Curve (ERC) \cite{hernandez2019faceqnet, boutros2021crfiqa,ou2021sdd_fiqa}, where portions of low-quality face images are screened out with respect to quality indexes. This evaluation is assessed in terms of False Non-Match Rate (FNMR) \cite{kukula2010human} at a specific threshold for a fixed False Match Rate (FMR) \cite{kukula2010human}.

\subsection{Comparison with SoTA Methods}
We compare the generalization performance on two open-set face identification tasks, VLR-VLR (TinyFace and SurvFace) and HR-VLR (SCFace), to the most recent SoTAs in \cref{table::TinyfaceSotaCompare},  \ref{table::scfaceSotaCompare}, and \ref{table::survfaceSotaCompare}\footnote{In the ``fine-tuned'' column of these tables, we indicate $\sharp$, $\dagger$, and $\checkmark$ as fine-tuning on super-resolved, synthetic down-sampled VLR, and native VLR face images respectively.}. The former task is relatively challenging as only the noisy VLR examples are provided for probe-to-gallery matching. However, the latter suffers from a severe resolution gap between the HR galleries and the VLR probes. We disclose that the proposed model is not only resistant to the resolution gap but also a viable solution to the downstream VLR-VLR task. On the other hand, we leave the column blank for KD and resolution-invariant methods without involving the VLR datasets for fine-tuning, seeing that these VLR datasets are relatively small-scale and therefore easily prone to over-fitting.

\newcolumntype{L}[1]{>{\raggedright\let\newline\\\arraybackslash\hspace{0pt}}m{#1}}
\newcolumntype{C}[1]{>{\centering\let\newline\\\arraybackslash\hspace{0pt}}m{#1}}

\begin{table*}
\begin{minipage}{0.49\textwidth}
\vspace{-0.5mm}
\renewcommand{\arraystretch}{1.263}
\centering
\resizebox{1.0\linewidth}{!}{\begin{tabular}{L{5.25cm} C{1cm} c c }
        \toprule
        \multirow{2}{*}{ Methods} & \multirow{2}{*}{Fine-} & \multicolumn{2}{c}{ Rank-1 IR (\%)}\\ \cmidrule(lr){3-4}
        {} & {Tuned} & { w/ dis} & { w/o dis}\\ \midrule
        { CSRI (ACCV19) \cite{cheng2018low_tinyface}} & {$\sharp$} & { 44.80} & {-} \\ 
        { TURL (CVPR20) \cite{shi2020URL}} & {} & { 63.89} & {-} \\
        { RIFR (T-BIOM20) \cite{khalid2020resolution}} & {} & { 70.40} & {-} \\ 
        { VividGAN (TIP21) \cite{zhang2021faceHalluwiffinishingtouches}} & {$\sharp$} & { 47.16} & {-} \\
        { MIND-Net (SPL21) \cite{low2021mindnet}} & {\checkmark} & { 66.82} & { 73.52} \\
        { AdaFace (CVPR22) \cite{kim2022adaface}} & {} & { 68.21} & - \\ 
        { IDEA-Net (TIFS22) \cite{low2022implicit}} & {\checkmark} & {68.13} & - \\ \hline
        { AdaFace (Reproduce) } & {\checkmark} & { 71.38} & {75.67} \\
        { Ours} & {\checkmark} & \textbf{ 73.06} & \textbf{ 77.22}\\
        \bottomrule
        \end{tabular}}
    \vspace{-2.5mm}
    \caption{IR Comparison on TinyFace using \textbf{ResNet-50}}
    \label{table::TinyfaceSotaCompare}
\end{minipage}\hfill 
\begin{minipage}{0.48\textwidth}
\renewcommand{\arraystretch}{1.0}
    \centering
\resizebox{\linewidth}{!}{\begin{tabular}{L{4.2cm} c c c c c  }
        \toprule
        \multirow{2}{*}{ Methods} & \multirow{2}{*}{ Fine-} & \multicolumn{4}{c}{ Rank-1 IR (\%)}\\  \cmidrule(lr){3-6}
        {} & {Tuned} & { 4.2m} & { 2.6m} & { 1.0m} & { Avg.}\\ \midrule
        { TCN (ICASSP10) \cite{zha2019tcn}} & {\checkmark}  & { 74.60} & { 94.90} & { 98.60} & { 89.37} \\
        { T-C (IVC20) \cite{massoli2020crossresolutionKD}} & {$\dagger$}  & { 70.20} & { 93.70} & { 98.10} & { 87.33} \\
        { FAN (ACCV19) \cite{yin2020FAN}} & {\checkmark}  & { 77.50} & { 95.00} & { 98.30} & { 90.30} \\
        { RAN (ECCV20) \cite{fang2020generate}} & {\checkmark}   & { 81.30} & { 97.80} & { 98.80} & { 92.63} \\
        { DDL (ECCV20) \cite{huang2020DDL}} & {\checkmark}  & { 86.80} & { 98.30} & { 98.30} & { 94.40} \\
        { RIFR (T-BIOM20) \cite{khalid2020resolution}} & {}  & { 88.30} & { 98.30} & { 98.60} & { 95.00} \\
        { MIND-Net (SPL21) \cite{low2021mindnet}} & {\checkmark}  & { 81.75} & { 98.00} & { 99.25} & { 93.00} \\
        { DSN (APSIPA21) \cite{lai2021deep}} & {\checkmark}  & { 93.00} & { 98.50} & { 98.50} & { 96.70} \\
        { DRVNet (TPAMI21) \cite{singh2021derivenet}} & {\checkmark}  & { 76.80} & { 92.80} & { 97.50} & { 89.03}\\
        { RPCL (NN22) \cite{li2022deeprival}} & {\checkmark}  & { 90.40} & { 98.00} & { 98.00} & { 95.46} \\
        { NPT (TPAMI22) \cite{khalid2022npt}} & {}  & { 85.69} & { 99.08} & { 99.08} & { 96.61} \\ 
        { IDEA-Net (TIFS22) \cite{low2022implicit}} & {\checkmark}  & {90.76} & {98.50} & {99.25} & {96.17} \\ \hline
        {AdaFace (Reproduce)} & {\checkmark}  & {95.38} & {98.46} & \textbf{99.84} & {97.89} \\
        { Ours} & {\checkmark}  & \textbf{ 97.07} & \textbf{ 99.23} & { 99.80} & \textbf{ 98.70} \\ 
        \bottomrule
\end{tabular}}
    \vspace{-2mm}
    \caption{IR Comparison on SCFace using \textbf{ResNet-50}}
    \label{table::scfaceSotaCompare}
  \end{minipage}
  \vspace{-1mm}
\end{table*}

\begin{table*}[t]
\renewcommand{\arraystretch}{1.0}
\centering
\resizebox{1.0\textwidth}{!}{\begin{tabular}{L{6cm} c C{1.3cm}  C{1.3cm}  C{1.3cm}  C{1.3cm}  C{1.3cm}  C{1.3cm}  C{1.3cm} }
\toprule
\multirow{2}{*}{ Methods} & \multirow{2}{*}{ Fine-} & \multicolumn{4}{c}{ TPR(\%)@FAR} & \multicolumn{3}{c}{ TPIR20(\%)@FPIR}\\ \cmidrule(lr){3-6} \cmidrule(lr){7-9}
{} & {Tuned} &  0.3 &  0.1 &  0.01 &  0.001 &  0.3 &  0.2 &  0.1\\ \hline
{ CSRI (ACCV19) \cite{jiao2021ddat, cheng2018low_tinyface}} & {$\sharp$} &  78.60 &  53.10 &  18.09 &  12.04 &  - &  - &  -  \\ 
{ FAN (ACCV19) \cite{yin2020FAN}} & {} &  71.30 &  44.59 &  12.94 &  2.75 &  - &  - &  -  \\ 
{ RAN (ECCV20) \cite{fang2020generate}} & {} &  - &  - &  - &  - &  26.50 &  21.60 &  14.90 \\ 
{ SST (ECCV20) \cite{du2020semi}} & {\checkmark} &  87.00 &  68.21 &  35.72 &  22.18 &  12.38 &  9.71 &  6.61 \\
{ DSN (APSIPA21) \cite{lai2021deep}} & {} &  75.09 &  52.74 &  21.41 &  11.02 &  - &  - &  -  \\ 
{ DDAT (PR21) \cite{jiao2021ddat}} & {} & \textbf{ 90.40} &  75.50 &  40.40 &  16.40 &  - &  - &  -  \\ 
{ IDEA-Net (TIFS22) \cite{fang2020generate}} & {} &  - &  - &  - &  - &  26.24 &  21.82 &  15.61 \\ \hline
{ AdaFace (Reproduce)} & {\checkmark} &  {87.41} & { 77.48} & { 58.63} & { 40.09} & { 31.50} & { 27.74} & { 21.93} \\ 
{ Ours} & {\checkmark} &  90.21 & \textbf{ 80.99} & \textbf{ 64.60} & \textbf{ 48.48} & \textbf{ 33.20} & \textbf{ 29.34} & \textbf{ 22.81} \\
\bottomrule
\end{tabular}}
\vspace{-2mm}
\caption{{TPR(\%)@FAR and TPIR20(\%)@FPIR Comparison} on SurvFace using \textbf{ResNet-50}}
\label{table::survfaceSotaCompare}
\vspace{-2mm}
\end{table*}


\textbf{TinyFace.} We substantiate that our model outperforms other recent SoTAs designated for VLRFR by a remarkable margin, both with and without distractors (a summation of 153,428 face images of unknown identities in the gallery set). Specifically, we demonstrate that improving recognizability at the feature level is more meaningful than super-resolving the visual quality of the VLR face images, e.g. \cite{cheng2018low_tinyface, zhang2021faceHalluwiffinishingtouches,shi2020URL,khalid2020resolution}. With a quality-adapted margin for embedding learning, we discern that AdaFace \cite{kim2022adaface} only readjusts the decision boundary - remaining the feature recognizability unchanged.

\textbf{SCFace}. For the HR-VLR evaluation, our model remains superior to other SoTAs over the three probe sets. This is attributed to: (i) the enhanced recognizability bridges the resolution gap between the HR and the VLR face features; and (ii) the attention module singles out the most salient regions from VLR and HR faces, resulting in better cross-resolution matching scores, especially for the face images captured from 4.2\textit{m} (the largest standoff distances compared to 1.0\textit{m} and 2.6\textit{m}).

\textbf{SurvFace.} Being the most challenging VLR face dataset for the VLR-VLR deployment scenario, SurvFace evaluates both open-set identification and verification tasks in the presence of 141,736 unmated distractors as a part of the probe set. We disclose that our model significantly outperforms other SoTAs under the most rigorous settings, i.e., TPR@FAR=0.001 and TPIR20@FPIR=0.01. While \cite{yin2020FAN, fang2020generate,jiao2021ddat} are trained based on synthetic LR images downsampled from HR-paired counterparts for visual quality improvement, we underline that none of these SoTAs resolves the VLRFR by means of refining feature recognizability.

\subsection{Ablation Analysis}
\vskip -0.5em
\textbf{Effect of Each Loss Component}. \textcolor{black}{
We present in \cref{table::ablation_loss} an ablation analysis to explore the effect of each loss term using MobileFaceNet on TinyFace without gallery distractors. Compare to our baseline model (trained using only ArcFace), the inclusion of ID loss in Baseline I discloses that learning to enhance the recognizability of the hard-to-recognize instances offers a performance improvement close to 1.0$\%$. Baseline II, on the other hand, reveals that the perceptibility attention module allows the most salient characteristics to be attended, resulting in at least 1.2$\%$ of performance gain. \textcolor{black}{Meanwhile, Baseline III demonstrates that our model benefits from learning an RI based on an estimated RI as shown in \cref{eq::smoothL1}. It is believed that learning the softmax prototypes simultaneously with the RI prompts our model to encode the embedding recognizability at each training step. As a result, the learned RI can be viewed as an index of the model's confidence corresponding to the classifiability of any face image, including the unknown instances.} Overall, our model trained with all loss terms outperforms the baselines and two most relevant SoTAs, i.e., MagFace \cite{meng2021magface}, and AdaFace \cite{kim2022adaface}. An important reason is that these SoTAs are reliant on the embedding's norm that does not always convey face recognizability, particularly for the VLR face images. This is substantiated by our empirical proofs in the following section. 
}

\begin{figure}[t]
    \centering
    \includegraphics[width=1.\linewidth]{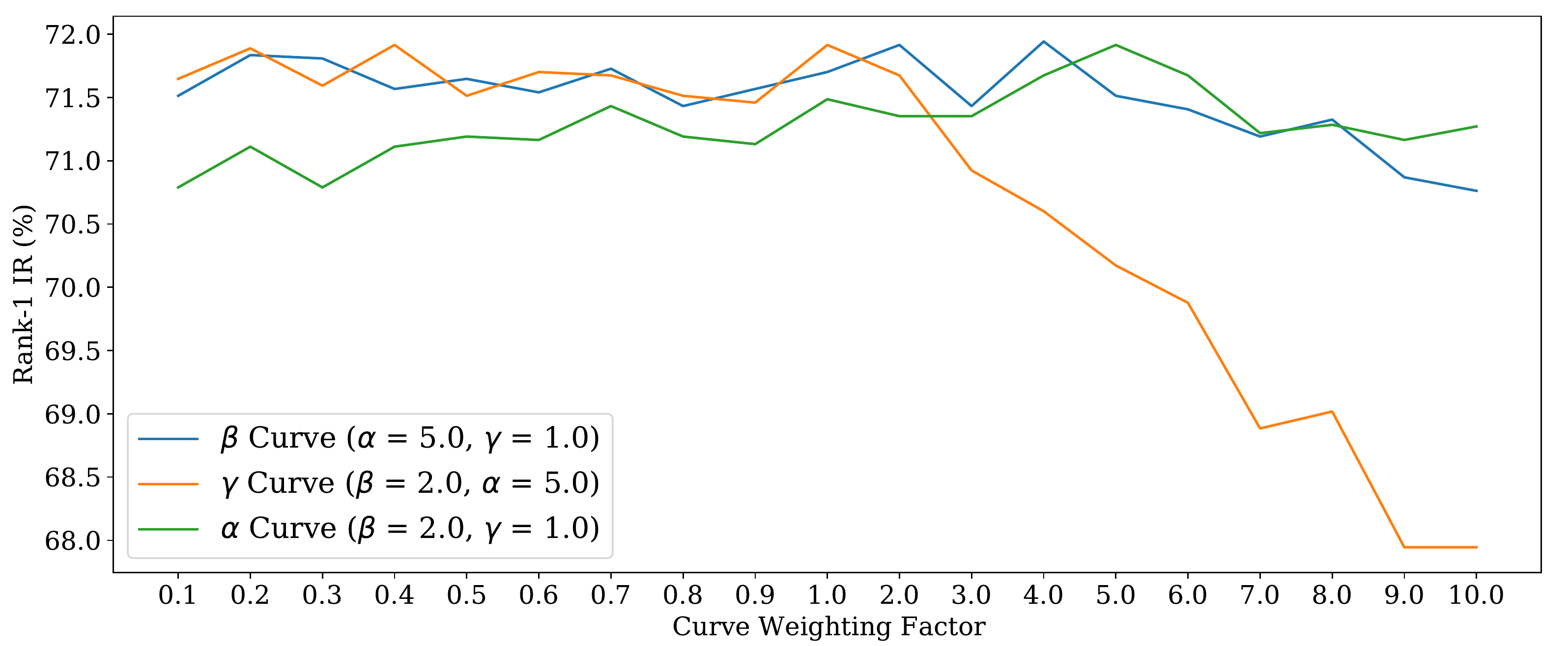}
    \vspace{-6mm}
    \caption{Ablation Studies for Hyperparameters $\alpha, \beta,$ and $\gamma$.}
    \label{fig::hyperparameters_ablation}
    \vspace{-2mm}
\end{figure}

\textbf{Intuition of Hyperparameter Selection}. In \cref{fig::hyperparameters_ablation}, we conduct the study of each weighting factor on each hyperparameter by fixing the remaining weighting factors to be their optimal values in accordance to \cref{table::hyperparameters}. We refer to \cref{eq::overall_euqation} where $\alpha, \beta, \gamma$ correspond to the weighting of $L_{L1}, L_{ID}, L_{MSE}$ respectively. Suppose that we attempt to study the various weighting factors of $L_{L1}$ via manipulating the values of $\alpha$, we fix $\gamma = 1$ and $\beta = 2$. The effects of $\alpha$ are shown as the $\alpha$ curve, and the same is applied to $\beta$ and $\gamma$ curves. Our ablations show that an overly large $\alpha (>10)$ may experience difficulty in convergence \cite{boutros2021crfiqa}, and 5 is the best choice. For $\beta$, the findings suggest a weighting factor slightly less than the $\alpha$ ensures better recognizability of the instances while still being upper bound by the devised RI. Lastly, we discover that $L_{MSE}$ converges fast, and a higher $\gamma$ hinders the overall model convergence, resulting in performance depreciation. A smaller $\gamma$ is thus recommended and set to be 1 in our experiment.

\begin{table}[t]
\renewcommand{\arraystretch}{1.0}
\centering
\resizebox{\linewidth}{!}{
\begin{tabular}{C{3.9cm} c c c c c }
\toprule
{{ Method}} & { $\mathcal{L}_{cls}$} & { $\mathcal{L}_{ID}$} & { $\mathcal{L}_{MSE}$} & { $\mathcal{L}_{L1}$} & { Rank-1 IR (\%)}\\ \cmidrule(lr){1-1}\cmidrule(lr){2-5}\cmidrule(lr){6-6}
{ Cross Entropy \cite{sun2014crossentropy}} & { $\checkmark$} & {} & {} & {} & { 68.884} \\
{ NormFace \cite{wang2017normface}} & { $\checkmark$} & {} & {} & {} & { 68.026} \\
{ CosFace \cite{wang2018cosface}} & { $\checkmark$} & {} & {} & {} & { 70.306} \\
{ MV-Softmax \cite{wang2020mis}} & { $\checkmark$} & {} & {} & {} & { 70.547} \\ 
{ CurricularFace \cite{huang2020curricularface}} & { $\checkmark$} & {} & {} & {} & { 70.655} \\ 
{ MagFace \cite{meng2021magface}} & { $\checkmark$} & {} & {} & {} & { 70.467} \\
{ AdaFace \cite{kim2022adaface}} & { $\checkmark$} & {} & {} & {} & { 70.359} \\ \hline
{ Baseline (ArcFace) \cite{deng2019arcface}} & { $\checkmark$} & {} & {} & {} & { 70.333} \\
{ I} & { $\checkmark$} & { $\checkmark$} & {} & {} & { 71.298} \\
{ II} & { $\checkmark$} & {} & { $\checkmark$} & {} & { 71.540} \\
{ III} & { $\checkmark$} & {} & {} & { $\checkmark$} & { 71.674} \\
{ Ours} & { $\checkmark$} & { $\checkmark$} & { $\checkmark$} & { $\checkmark$} & \textbf{ 71.915} \\
\bottomrule
\end{tabular}}
\vspace{-2.5mm}
\caption{Ablation Analysis for each Loss Term on TinyFace \textit{without Distractors} using \textbf{MobileFaceNet}.}\label{table::ablation_loss}
\vspace{-1mm}
\end{table}

\begin{figure}[t!]
\centering
\includegraphics[width=0.99\linewidth]{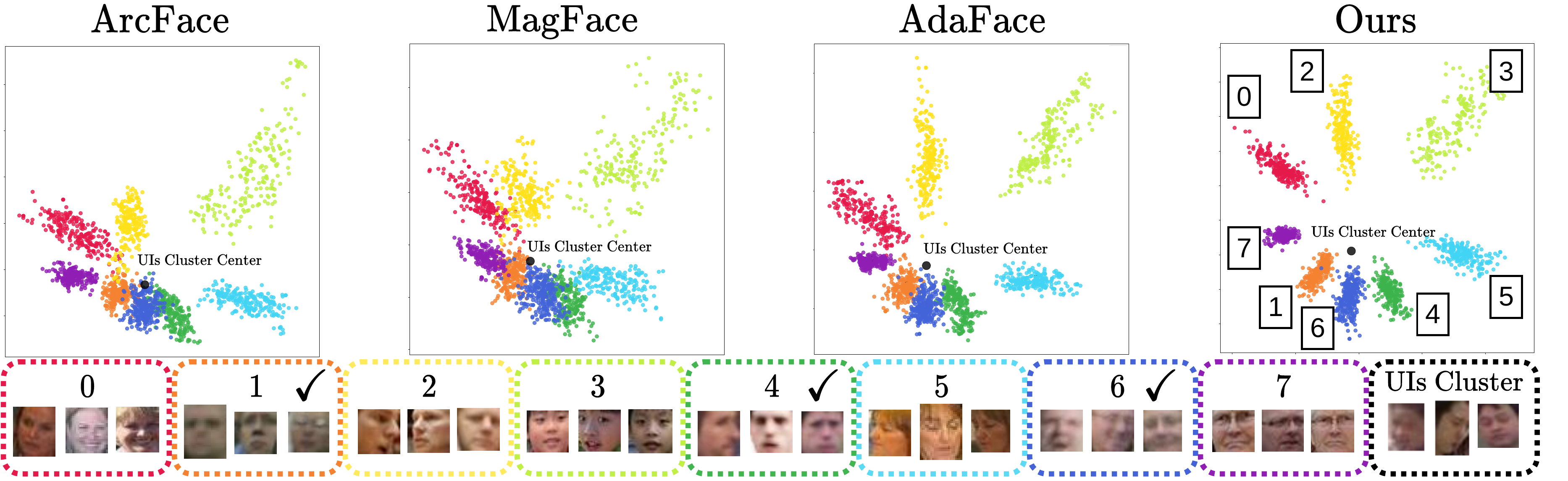}
    \vspace{-2mm}
    \caption{2D embedding space for a toy problem with 8 identities (IDs). The IDs denoted with a "$\checkmark$" contain hard-to-recognize instances, even from the human perspective. We visualize the UIs center as a reference index of recognizability, i.e., embeddings further away from the UIs center can better be recognized.}
    \label{fig::2d_flower}
    \vspace{-3mm}
\end{figure}

\textbf{Face Recognizability-Aware Embeddings.} We conduct a toy experiment with 5 easy-to-recognize (IDs 0,2,3,5,7) and 3 hard-to-recognize (IDs 1,4,6) face examples to examine the embedding space learned by SoTAs and our model in \cref{fig::2d_flower}. The UIs center is visualized as a reference index of recognizability, where a poor recognizability (hard-to-recognize) instance is essentially projected close to the UIs center. It is discerned that the competing models are incapable of separating the hard-to-recognize clusters from the UIs center. On the contrary, our model is inclined to divert the hard-to-recognize instances from the UIs center, yielding a well-separable (therefore a more discriminative) embedding space to resolve the downstream VLRFR task.

\begin{figure}[t!]
\centering
\includegraphics[width=0.95\linewidth]{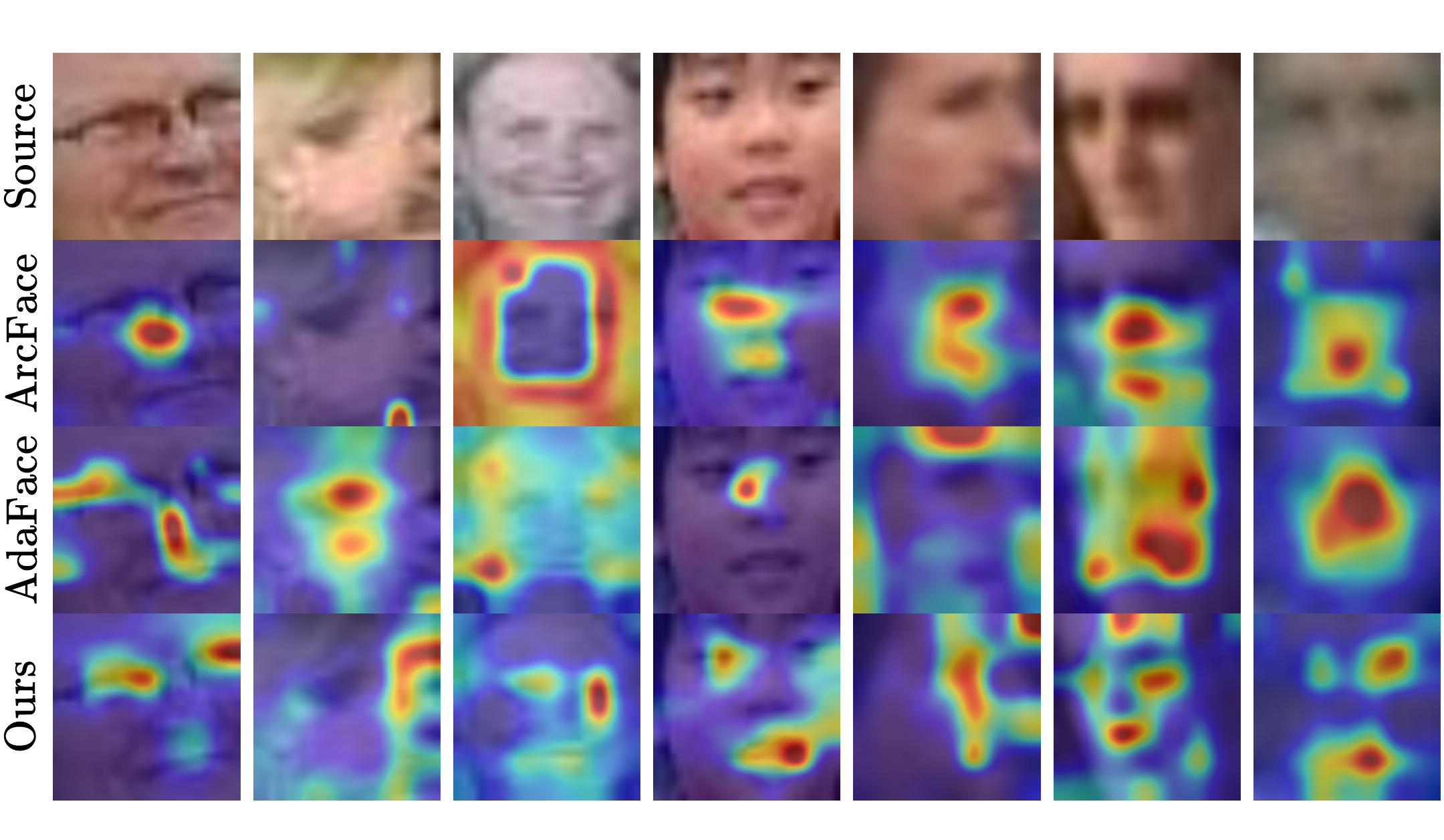}
    \vspace{-2mm}
    \caption{Class activation maps for several VLR face images generated based on ArcFace, AdaFace, and Ours.}
    \label{fig::heatmap}
    \vspace{-2mm}
\end{figure}

\textbf{Effect of Perceptibility Attention Module.} We portray the Class Activation Maps \cite{selvaraju2017gradcam} obtained from ArcFace \cite{deng2019arcface}, Adaface \cite{kim2022adaface}, and our method for comparison. Interestingly, we observe from \cref{fig::heatmap} that attending to the embeddings away from the UIs center focuses more on the salient face features, i.e. eyes, nose, and mouth, even when the face images appear to be obscure. We provide the extended heatmap visualizations in \cref{fig::heatmap_extension} of supplemental material.

\subsection{Face Image Quality Assessment (FIQA)}
\begin{figure*}[t!]
\centering
\includegraphics[width=0.99\linewidth]{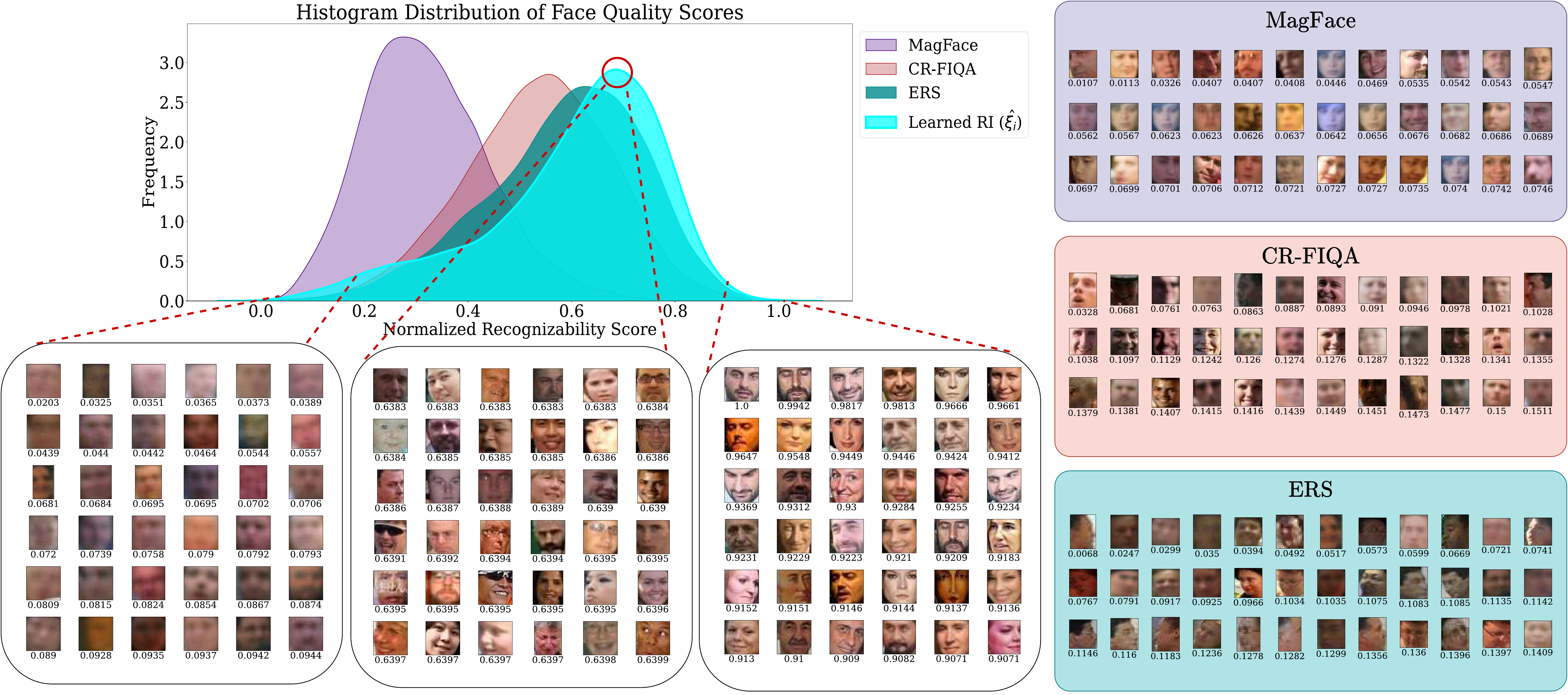}
    \vspace{-2mm}
    \caption{To facilitate direct comparison, all quality measures are normalized to ranges between 0 and 1. (\textbf{Left}) Visualization of RI sorted from the lowest scores (the poorest quality) to the highest (the best quality) in TinyFace \cite{cheng2018low_tinyface} dataset. It is disclosed that the proposed RI characterizes the face quality better, closely simulating the human cognitive level. (\textbf{Right}) The VLR face images estimated with the lowest quality scores based on MagFace \cite{meng2021magface}, CR-FIQA \cite{boutros2021crfiqa} and ERS \cite{deng2021harnessing}. In place of hard-to-recognize instances, we observe that several high-quality face instances are mistakenly assigned with low-quality scores, particularly MagFace and CR-FIQA. }
    \label{fig::histogram_visualization_with_face_examples}
    \vspace{-1mm}
\end{figure*}

\begin{figure*}[t]
\centering
    \includegraphics[width=0.99\linewidth]{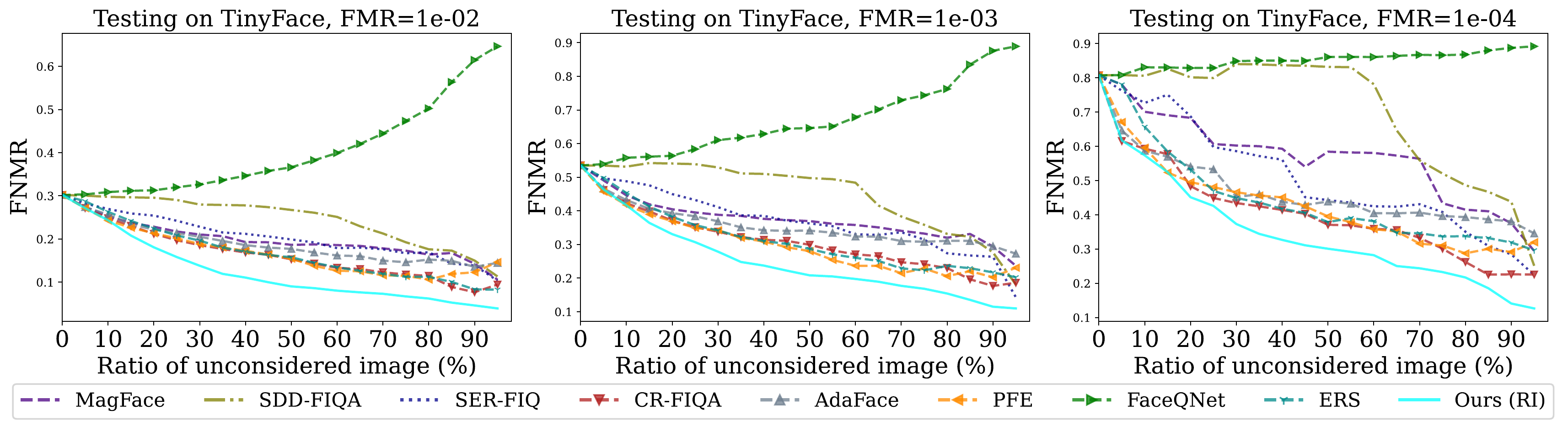}
    \vspace{-2mm}
    \caption{
    Whilst \cref{fig::histogram_visualization_with_face_examples} shows that the learned RI can characterize face quality well, we further demonstrate in this figure that the learned RI is a reliable recognizability metric by means of analyzing false non-match rate (FNMR). In particular, FNMR decreases gradually when the ratio of unconsidered VLR face images increases (sampled based on the lowest RI). This indicates that the learn RI well characterizes the recognizability of the VLR face images, such that the most recognizable VLR face images are learned with the highest RIs (corresponding to the right of the left-skewed RI distribution) for FNMR evaluation.
    }
    \label{fig::ERC_Curve}
    \vspace{-2mm}
\end{figure*}

\textbf{Learned RI from Perceptibility Regression Module}. In \cref{fig::histogram_visualization_with_face_examples}, the effect of ID loss reveals a greater negatively-skewed distribution of RI than other SoTAs (extended version in \cref{fig::all_histogram_extended}), indicating an improvement in recognizability, notably the hard-to-recognize instances. We discern that the proposed RI is a robust indicator in characterizing the recognizability notion - the highest and the lowest RI reflect the best-quality and the poor-quality face images, respectively. We also provide in \cref{fig::histogram_visualization_with_face_examples} the face images with the lowest quality scores estimated by other competing scores. Through the comparison, RI is deemed proportional to the human cognitive level in terms of recognizability for the VLR face images under extreme poses, illuminations, occlusions, and others in the wild conditions. 

\textbf{Error versus Rejection Curve (ERC)}. To further substantiate the RI's robustness, we perform a verification task on TinyFace \cite{cheng2018low_tinyface} by randomly sampling 20,888 positive and 50,000 negative pairs from its gallery and probe sets to show ERC in \cref{fig::ERC_Curve}. The learned $\widehat{\xi}_i$ outperforms SoTA in achieving stable and true rejection of unrecognized pairs in the rank of recognizability, especially when the FMR is at $1e^{-4}$. 

\vspace{-1.5mm}
\section{Conclusion}
\vspace{-1mm}
\label{section::conclusion}
This paper addresses the problem arising from the hard-to-recognize faces in VLR images. Rather than treating these faces as UIs, we take a principled recognizability notion to characterize the recognizability of each image with a robust indicator, specifically the recognizability index (RI). The recognizability of an instance can thus be adjusted based on RI. Interestingly, attending to the embeddings projected away from the UIs cluster provides more explanatory power to the model to highlight the facial features more precisely. We evaluate the proposed method trained in an end-to-end manner on three VLR datasets and achieve SoTA for both VLRFR and FIQA.

\vspace{-1.5mm}
\section*{Acknowledgements}
\vspace{-1mm}
This work was supported by the National Research Foundation of Korea (NRF) grant funded by the Korea government (MSIP) (NO. NRF-2022R1A2C1010710) and Institute for Basic Science (IBS-R029-C2) Korea.

{\small
\bibliographystyle{ieee_fullname}
\bibliography{ref}
}

\clearpage

\setcounter{section}{0}
\renewcommand*{\thesection}{\Alph{section}}
\setcounter{page}{1}
\renewcommand{\thepage}{\arabic{page}}

\section*{Supplemental Materials}

\section{Realistic Low-Resolution Datasets}
\label{sec::dataset_description}
We elaborate the face datasets involved in our experiments in this section. These include three realistic VLR datasets for benchmarking under the open-set evaluation protocol, i.e. Tinyface, SCface, and SurvFace, alongside a UI face dataset for recognizability index (RI) learning.

\textbf{TinyFace}.
TinyFace \cite{cheng2018low_tinyface} is a composition of 7,804 and 8,171 VLR face images annotated with 2,570 and 2,569 identity labels in each training and testing set, respectively. On average, the input resolution for each VLR face image is of 20$\times$16 pixels. The gallery search space is interfered with 153,428 distractors of unknown identities to simulate a more challenging open-set VLR-VLR identification scenario.

\textbf{SurvFace}. 
SurvFace \cite{cheng2018survface} is the largest surveillance face dataset for VLR-VLR identification and verification tasks evaluated in the presence of unmated distractors (probe images without a matched gallery ID). In a nutshell, it contains 463,507 VLR face images contributed by 15,573 subjects with an average resolution of 24$\times$20 pixels. For the VLR-VLR identification task, it is partitioned with 220,890 face images for 5,319 subjects in the training set, whilst the testing set consists of 242,617 face images for 10,254 subjects (including 141,736 unmated probe images from 4,935 subjects). On the other hand, it is sampled with 5,319 matched and unmatched pairs for evaluating the VLR-VLR verification task.

\textbf{SCFace}. 
Unlike TinyFace and SurvFace, SCFace \cite{grgic2011scface} is a small-scale face dataset with HR and VLR face images for the HR-VLR identification task. Overall, each of the 130 subjects is provided with a single HR mugshot (as a gallery template), alongside 15 VLR face images captured from three standoff distances, i.e. 4.20$m$ (D1), 2.60$m$ (D2) and 1.00$m$ (D3). Our experiments allocate the face images for the first 50 subjects (from ID \textit{001} to \textit{050}) for training, excluding the corresponding HR gallery templates. The remaining subjects (from ID \textit{051} to \textit{130}) are probed with respect to all 130 gallery templates.

\begin{figure}[htp!]
    \centering
    \includegraphics[width=0.99\linewidth]{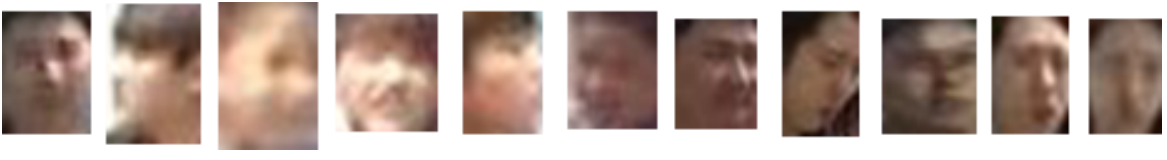}
    \caption{Sample face images in our UI dataset.}
    \label{fig::UI_dataset_examples}
\end{figure}

\textbf{UI Face Dataset}. To elicit an UIs cluster for RI learning, we assemble an ad-hoc UI face dataset with a summation of 11,707 unlabeled VLR face images. We single out these VLR face images from two person re-identification datasets (independent from the three benchmarking datasets), i.e. LPW \mbox{\cite{song2018lpwdataset}} and MARS \mbox{\cite{zheng2016mars}}, by the MTCNN face detector \mbox{\cite{zhang2016mtcnn}}. As illustrated in Fig.~\mbox{\ref{fig::UI_dataset_examples}}, these VLR face images are close to unrecognizable from the image quality perspective. 

\section{Experiment Settings}
\label{section::experiment_settings}
Given face images of low pixel resolutions in TinyFace, SurvFace, SCFace, and the UI dataset, we rescale all the images into 112$\times$112 pixels through bi-cubic interpolation. The rescaled images are forwarded to the pretrained MobileFaceNet \cite{chen2018mobilefacenets}, and ResNet-50 \cite{deng2019arcface} for embedding learning in the training stage, followed by feature extraction in the inference stage. Since the validation set is not available for all datasets for hyperparameter tuning, we sample a random subset with definite horizontal flip and extreme downsampled counterparts by 16$\times$16 pixels from the respective training sets. For consistent batch-wise statistics, we suspend the pre-trained batch-normalization (BN) layers from learning - applicable to all the BN layers after every 2D convolutional layer. Our model is trained with an Adam optimizer and augmented instances, including random horizontal flip, random rotation (in the range of ±10 degree), and rescaling (by 64$\times$64, 100$\times$100). Our implementation is exercised using a machine with two NVIDIA 2080Ti GPUs in the PyTorch framework.

In \cref{table::hyperparameters}, we provide a complete description of the ResNet-50 hyperparameters used for all three datasets defined in \cref{sec::dataset_description}, namely TinyFace, Survface, and SCFace. We refer all modules mentioned in \cref{table::hyperparameters} to \cref{fig::arch} to aid the readability. For MobileFaceNet, the learning rate for backbone, classifier head and perceptibility attention module are changed to $1e^{-4}$, $1e^{-3}$ and $1e^{-2}$ respectively. The others remain the same unless otherwise specified. These hyperparameters are determined based on the validation sets sampled from the corresponding training set.

\begin{table*}[ht!]
\centering
\resizebox{\textwidth}{!}{
\begin{threeparttable}
\begin{tabular}{l| l r l| l r  l| l r }
\toprule
\multicolumn{3}{c}{TinyFace} & \multicolumn{3}{c}{SCFace} & \multicolumn{3}{c}{Survface}\\ \cmidrule(lr){1-3} \cmidrule(lr){4-6} \cmidrule(lr){7-9}

\multicolumn{2}{l}{Mini-Batch Size} & {64} & \multicolumn{2}{l}{Mini-Batch Size} & {64} & \multicolumn{2}{l}{Mini-Batch Size} & {100} \\
\multicolumn{2}{l}{\# Epoch} & {15} & \multicolumn{2}{l}{\# Epoch} & {6} & \multicolumn{2}{l}{\# Epoch} & {30} \\
\multirow{4}{*}{\parbox{1.5cm}{Learning Rate}} & {Backbone} & {$1e^{-5}$} & \multirow{4}{*}{\parbox{1.5cm}{Learning Rate}} & {Backbone} & {$1e^{-5}$} & \multirow{4}{*}{\parbox{1.5cm}{Learning Rate}} & {Backbone} & {$1e^{-5}$} \\
{} & {RM\tnote{1}} & {$1e^{-3}$} & {} & {RM\tnote{1}} & {$1e^{-3}$} & {} & {RM\tnote{1}} & {$5e^{-4}$} \\
{} & {PAM\tnote{2}} & {$1e^{-3}$} & {} & {PAM\tnote{2}} & {$1e^{-3}$} & {} & {PAM\tnote{2}} & {$1e^{-3}$} \\
{} & {PRM\tnote{3}} & {$1e^{-4}$} & {} & {PRM\tnote{3}} & {$1e^{-4}$} & {} & {PRM\tnote{3}} & {$1e^{-4}$} \\

\multicolumn{2}{l}{Learning Rate Decay} & {0.1 / $12^{th}$ epoch} & \multicolumn{2}{l}{Learning Rate Decay} & {0.1 / $12^{th}$ epoch} & \multicolumn{2}{l}{Learning Rate Decay} & {0.1 / $6^{th}$ epoch} \\
\multicolumn{2}{l}{Dropout (Backbone)} & {0.2} & 
\multicolumn{2}{l}{Dropout (Backbone)} & {0.2} & 
\multicolumn{2}{l}{Dropout (Backbone)} & {0.4} \\
\multicolumn{2}{l}{Dropout (PRM\tnote{3} \,)} & {0.9} & 
\multicolumn{2}{l}{Dropout (PRM\tnote{3} \,)} & {0.9} & 
\multicolumn{2}{l}{Dropout (PRM\tnote{3} \,)} & {0.9} \\
\multicolumn{2}{l}{Weight Decay} & {$1e^{-4}$} & \multicolumn{2}{l}{Weight Decay} & {$1e^{-4}$} & \multicolumn{2}{l}{Weight Decay} & {$1e^{-3}$} \\
\multicolumn{2}{l}{$s, m$} & {64, 0.45} & \multicolumn{2}{l}{$s, m$} & {64, 0.45} & \multicolumn{2}{l}{$s, m$} & {64, 0.45} \\
\multicolumn{2}{l}{$\alpha, \beta, \gamma$} & {5, 2, 1} & \multicolumn{2}{l}{$\alpha, \beta, \gamma$} & {5, 2, 1} & \multicolumn{2}{l}{$\alpha, \beta, \gamma$} & {5, 2, 1} \\
\bottomrule
\end{tabular}
\begin{tablenotes}
\item [1] Recognition Module
\item [2] Perceptibility Attention Module
\item [3] Perceptibility Regression Module
\end{tablenotes}
\caption{Hyperparameter Configuration for \textbf{ResNet-50}}
\label{table::hyperparameters}
\end{threeparttable}
}
\end{table*}

\section{Extended Visualization}
\label{sec::extended_vis_qualit_score_dis}
\begin{figure*}[htp!]
     \includegraphics[width=0.99\textwidth]{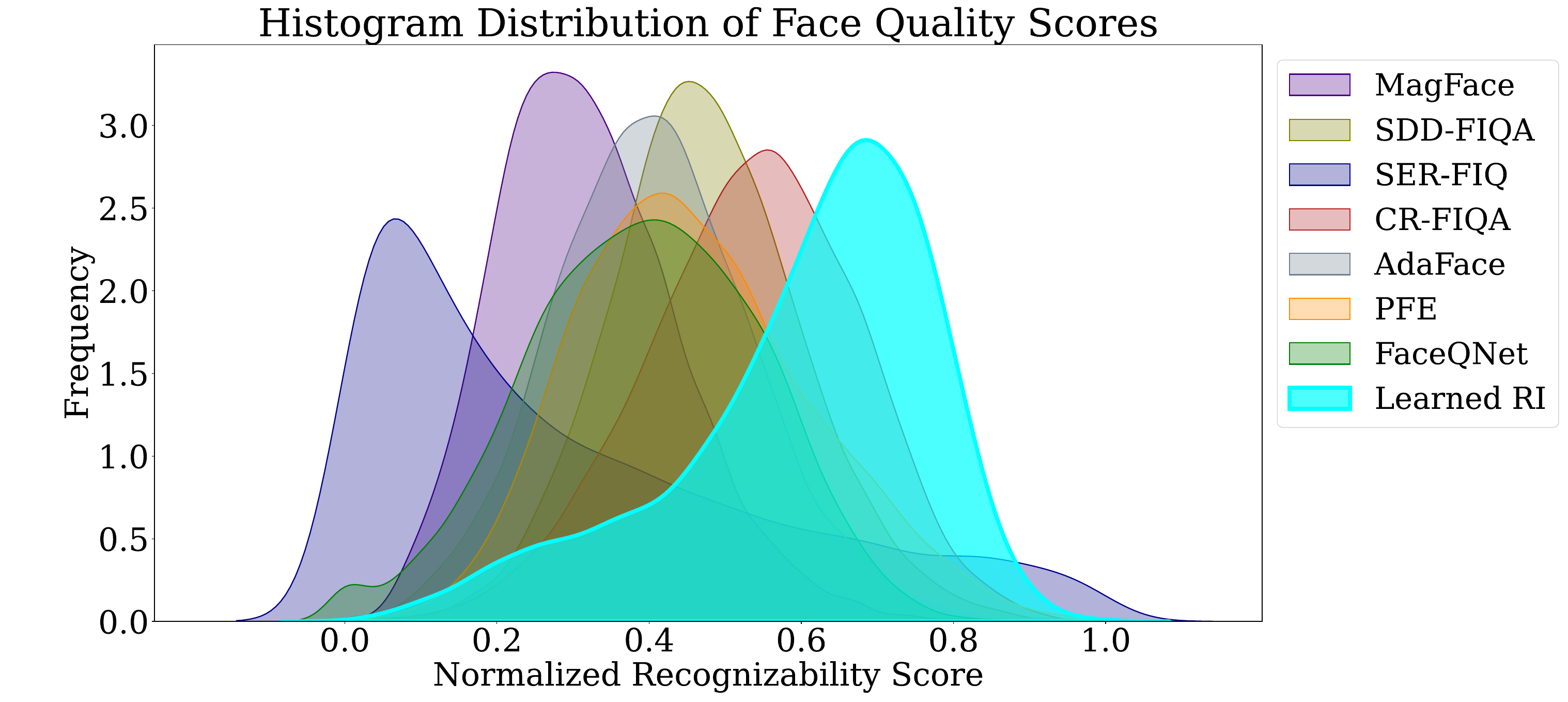}
    \caption{
    Quality score distributions for state-of-the-art metrics. In comparison, the score distribution for our learned RI is left-skewed. This indicates that the recognizability for majority of the VLR instances are enhanced, leaving minority of the hard-to-recognize VLR instances at lower scores. }
    \label{fig::all_histogram_extended}
\end{figure*}

An extended visualization of \cref{fig::histogram_visualization_with_face_examples} is shown in \cref{fig::all_histogram_extended}.

\begin{figure*}[htp!]
     \includegraphics[width=0.99\textwidth]{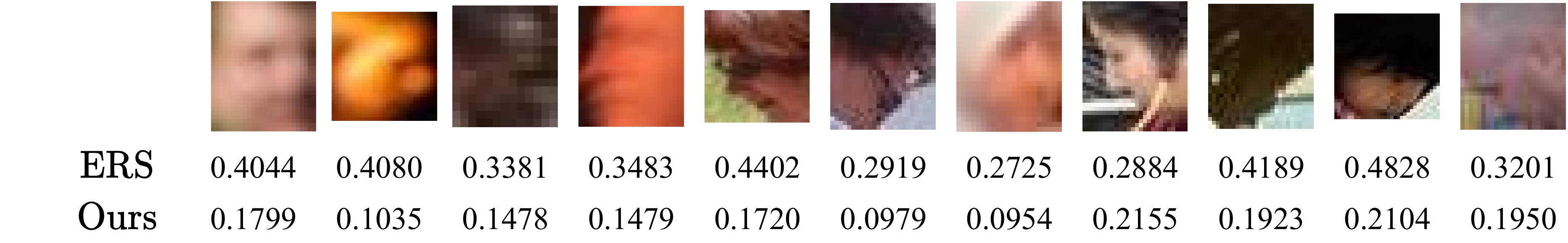}
    \caption{Comparison of recognizability between ERS \cite{deng2021harnessing} and our proposed RI for VLR face images captured under various unconstrained surveillance scenarios. Note that the lower the value, the harder the VLR face image is to be recognized.}
    \label{fig::bottom_various_pose}
\end{figure*}

For further exploration, we examine the ERS scores and RI with additional hard-to-recognize instances in \cref{fig::bottom_various_pose}. We demonstrate that RI is proportional to the human cognitive level over ERS for the VLR face images with extreme poses, illuminations, occlusions, and others in the wild conditions.

\begin{figure*}[t]
    \includegraphics[width=0.99\textwidth]{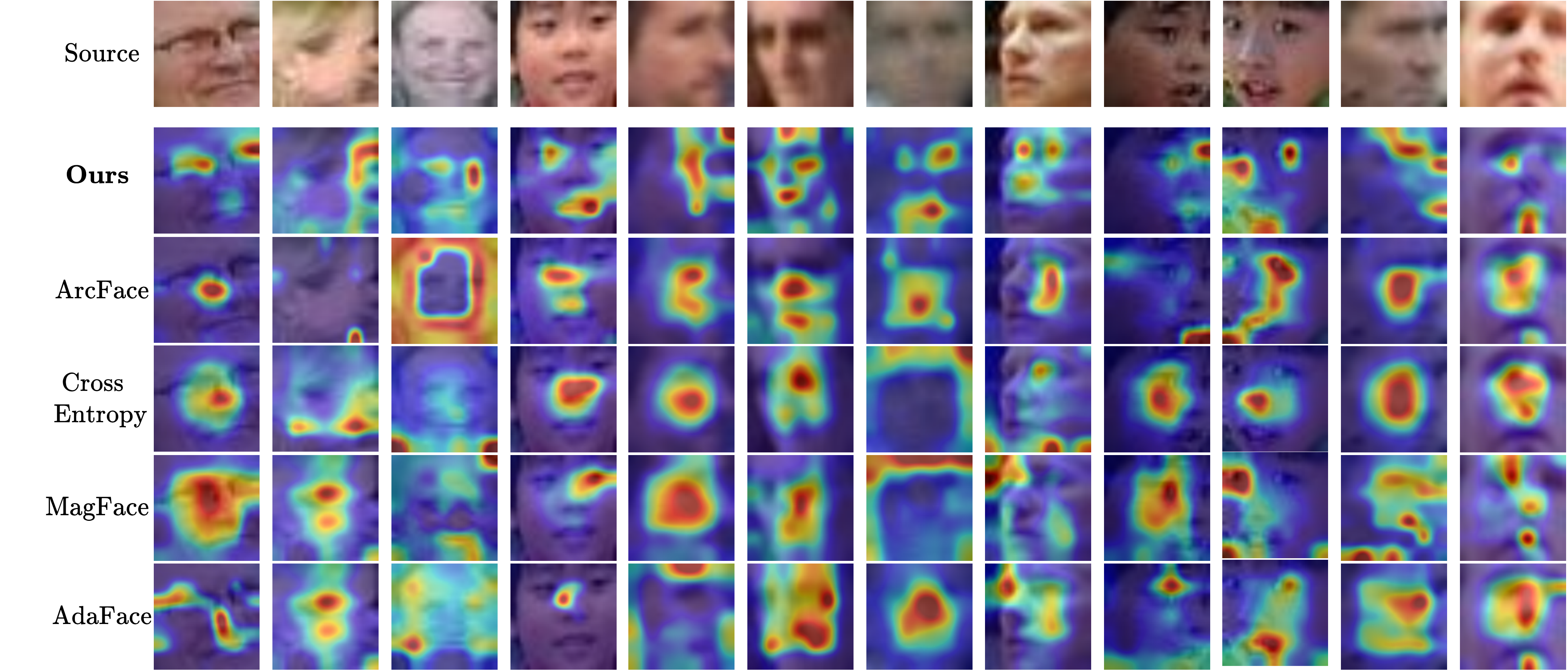}
    \caption{Class Activation Maps (extension to \cref{fig::heatmap}) for our RI, ArcFace \cite{deng2019arcface} (Baseline), Cross Entropy, MagFace \cite{meng2021magface}, and AdaFace \cite{kim2022adaface}.}
    \label{fig::heatmap_extension}
\vspace{50cm}
\end{figure*}

On top of \cref{fig::heatmap}, we also provide an extended visualization of class activation maps rendered based upon varying softmax-based losses in \cref{fig::heatmap_extension}, including the most recent SoTA - AdaFace \cite{kim2022adaface}. While some methods can capture certain parts of a face, our method provides more consistent attention to the salient facial features and is more capable of handling multiple poses, especially when the face is facing sideways or not aligned to the center of an image.

\end{document}